%% file: neurips_2024.tex
\documentclass{article}

\usepackage[nonatbib,final]{neurips_2024}

\usepackage[utf8]{inputenc} 
\usepackage[T1]{fontenc}    
\usepackage{hyperref}       
\usepackage{url}            
\usepackage{booktabs}       
\usepackage{amsfonts}       
\usepackage{nicefrac}       
\usepackage{microtype}      
\usepackage{xcolor}         

\usepackage{subfiles}
\usepackage[nottoc]{tocbibind}
\usepackage{textcomp}
\usepackage[square, numbers, sort]{natbib}
\usepackage{graphicx}
\usepackage{amssymb}
\usepackage{graphicx}
\usepackage{subfigure}
\usepackage{threeparttable}
\usepackage[marginal, flushmargin]{footmisc}
\usepackage{algorithm}
\usepackage{algorithmic}
\usepackage[algo2e,ruled,linesnumbered]{algorithm2e}
\usepackage{multirow}
\usepackage{bm}
\usepackage{amsmath}
\usepackage{amsthm}
\usepackage{makecell}

\usepackage{comment}
\usepackage{color}
\usepackage{etoolbox}
\newbool{inccomment}
\setbool{inccomment}{true}

\usepackage[utf8]{inputenc}
\usepackage[english]{babel}

\usepackage{wrapfig}
\usepackage{caption}
\usepackage[titletoc,title]{appendix}

\newcommand{\wc}[1]{\ifbool{inccomment}{{\color{blue}#1}}{}}
\newcommand{\ww}[1]{\ifbool{inccomment}{{\color{magenta} #1}}{}}
\newcommand{\rz}[1]{\ifbool{inccomment}{{\color{red} #1}}{}}

\title{CubicML: Automated ML for Large ML Systems Co-design with ML Prediction of Performance}

\author{%
  Wei Wen, Quanyu Zhu, Weiwei Chu, Wen-Yen Chen, Jiyan Yang \\
  AI at Meta\\
  \texttt{\{wewen, qyz, wchu, wychen, chocjy\}@meta.com} \\
}

\begin{document}

\maketitle

\begin{abstract}
  Scaling up deep learning models has been proven effective to improve intelligence of machine learning (ML) models, especially for industry recommendation models and large language models. The co-design of large distributed ML systems and algorithms (to maximize training performance) plays a pivotal role for its success. As it scales, the number of co-design hyper-parameters grows rapidly which brings challenges to feasibly find the optimal setup for system performance maximization. In this paper, we propose CubicML which uses ML to automatically optimize training performance of large distributed ML systems. In CubicML, we use an ML model as a proxy to predict the training performance for search efficiency and performance modeling flexibility. We proved that CubicML can effectively optimize training speed of in-house ads recommendation models with $73$ billion parameters and large language models up to 405 billion parameters at Meta.
\end{abstract}

\input{introduction}

\input{method}
\input{experiment}

\bibliography{neurips_2024}

\appendix
\input{appendix}

\end{document}

%% file: introduction.tex
\section{Introduction}

Scaling deep learning models and training data has become a standard solution to improve the intelligence of machine learning (ML) models, typically of generative AI models~\cite{kaplan2020scaling,peebles2023scalable} and industry-level recommendation models~\cite{zhang2024wukong,anil2022factory,ardalani2022understanding}.
To enable efficient large scale training, it is essential to co-design ML algorithms in distributed ML systems, such as data parallelism by ZeRO~\cite{rajbhandari2020zero,rasley2020deepspeed} and FSDP~\cite{zhao2023pytorch}, model parallelism in the forms of tensor parallelism and pipeline parallelism~\cite{shoeybi2019megatron,narayanan2021efficient}, low precision training~\cite{wen2017terngrad,peng2023fp8} and more.
While human experts are proficient at proposing these co-design algorithms, they face challenges to maximize training speed/performance by effectively selecting hyper-parameters of those co-design algorithms based on current model and system setup, such as, to maximize training performance under memory constraint by applying different layer-wise FSDP/ZeRO data parallelism strategies, to accelerate large language model (LLM) distributed training by selecting different parallelism strategies and their hyper-parameters according to the scale and architecture of LLM models, and so on. 
This problem becomes bigger when ML systems scale up with more co-design hyper-parameters. Finding the best hyper-parameter for ML system efficiency/performance is beyond easy reach of human experts by manual tuning. Moreover, as the model architecture and system hardware keep involving, repeated tuning is required whenever a change happens, demanding enormous human resources during development.
This difficulty calls for a more principled and automated approach to search co-design hyper-parameters.
Meanwhile, automated machine learning (AutoML~\cite{he2021automl}) has been successfully applied to search ML algorithms, such as model architecture and optimization hyper-parameters, providing one promising solution we can adopt for distributed ML system co-design; however, AutoML, typically black-box optimization, has surprisingly limited applications to distributed ML system co-design, likely because search space was small and simple grid search was enough.
In modern distributed ML systems with growing co-design search space, AutoML becomes non-trivial.
On the other hand, performance modeling of distributed ML systems is a key component to optimize system efficiency.
A performance model is usually lightweight and run efficiently, such that it can be used as a proxy to optimize systems.
However, to build a performance model, enormous research and engineering efforts in ML systems and ML algorithms are required, including ML algorithm deep dive, system hardware understanding, profiling, testing and refining. An accurate performance model is only within reach of experts with high expertise of both ML systems and ML algorithms; moreover, the performance model lacks generalization and a redesign is demanded whenever the hardware system or model algorithm changes.
Instead of manually customizing a performance model, we show that simple ML models can accurately predict performance of distributed ML training, even for a very complex search space in LLMs. Moreover, this method does not require deep dive into the details of current ML systems and adapts well because of its online learning ability.

There exist some work~\cite{zheng2022alpa,zhang2023auto,lin2023uniap,chen2018tvm} on auto-tuning system efficiency with performance modeling but they are highly tailored to  specific co-design situations (i.e. a specific parallelism~\cite{lin2023uniap} or ML compiler~\cite{chen2018tvm}) in a small system~\cite{zhang2023auto}.
We target on a generic black-box AutoML solution to optimize large-scale distributed ML systems by building ML prediction of system performance.
In this paper, we introduce CubicML, an automated machine learning algorithm for distributed ML system co-design to optimize training efficiency. CubicML is first comprehensively evaluated in large-scale ads recommendation models when optimizing layer-wise ZeRO sharding strategy; and then CubicML is evaluated in large language models (LLM) when searching model architecture, data and model parallelism strategies, training precision (FP8 and BF16) and others in different distributed system infrastructure setup (e.g. the number of GPUs and hardware types).
In recommendation models, we compared CubicML with baselines designed by in-house engineers by sampling real profiling jobs as dataset to build a ML model for performance prediction; in LLM models, we used hundreds of LLM training jobs to prove the accurate prediction of training performance by CubicML at scale.

%% file: method.tex
\section{The CubicML Framework}

\begin{figure}
\centering
\includegraphics[width=1.0\textwidth]{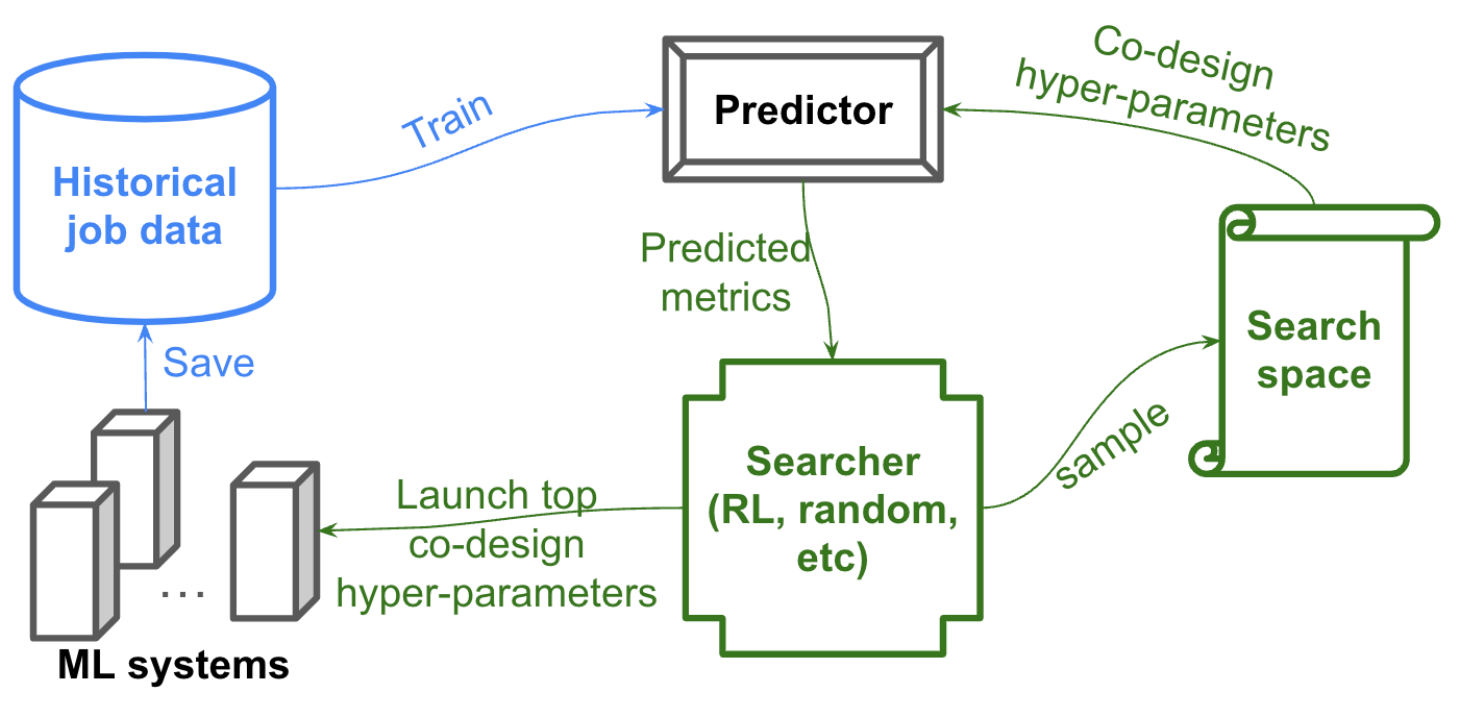}
\caption{CubicML framework overview.}
\label{fig:autoco}
\end{figure}

Figure~\ref{fig:autoco} illustrates CubicML, which adopts an algorithm close to~\cite{yin2024automl,wen2020neural}. There are five key components: \textbf{(1) ML systems} which is the training cluster stack to launch ML training jobs. It takes co-design hyper-parameters as job configurations to launch a specific ML training. When the job completes, its metadata (e.g. the hyper-parameters, training speed, etc) is saved;
\textbf{(2) historical job data} which stores the metadata of completed jobs. It is used as dataset to train a regression model (dubbed as ``predictor'') to predict ML metrics (such as training speed) for co-design hyper-parameters; 
\textbf{(3) search space} which defines a set of co-design hyper-parameters with their value ranges that CubicML can tune; 
\textbf{(4) predictor} is a lightweight regression model such as a neural network or decision tree regressor to predict system performance (i.e. training speed) we target to optimize. Margin Ranking Loss is used to train the predictor; 
\textbf{(5) searcher} which defines a search algorithm to sample many sets of hyper-parameters from the search space, feed these hyper-parameters to the predictor to predict corresponding system metrics, select hyper-parameters with top metrics to launch real training jobs into the ML systems. The searcher can be any black box optimizer, such as random, reinforcement learning (RL), Bayesian method, evolutionary algorithms, and so on. We use RL with REINFORCE algorithm in CubicML. We perform multiple rounds of RL search with different seeds to increase the diversity of top co-design hyper-parameters predicted by the ``predictor''.
    

%% file: experiment.tex
\section{Experiment}

\subsection{ZeRO Sharding Optimization for Distributed Training of Recommendation Models}

\begin{figure}
\centering
\includegraphics[width=0.9\textwidth]{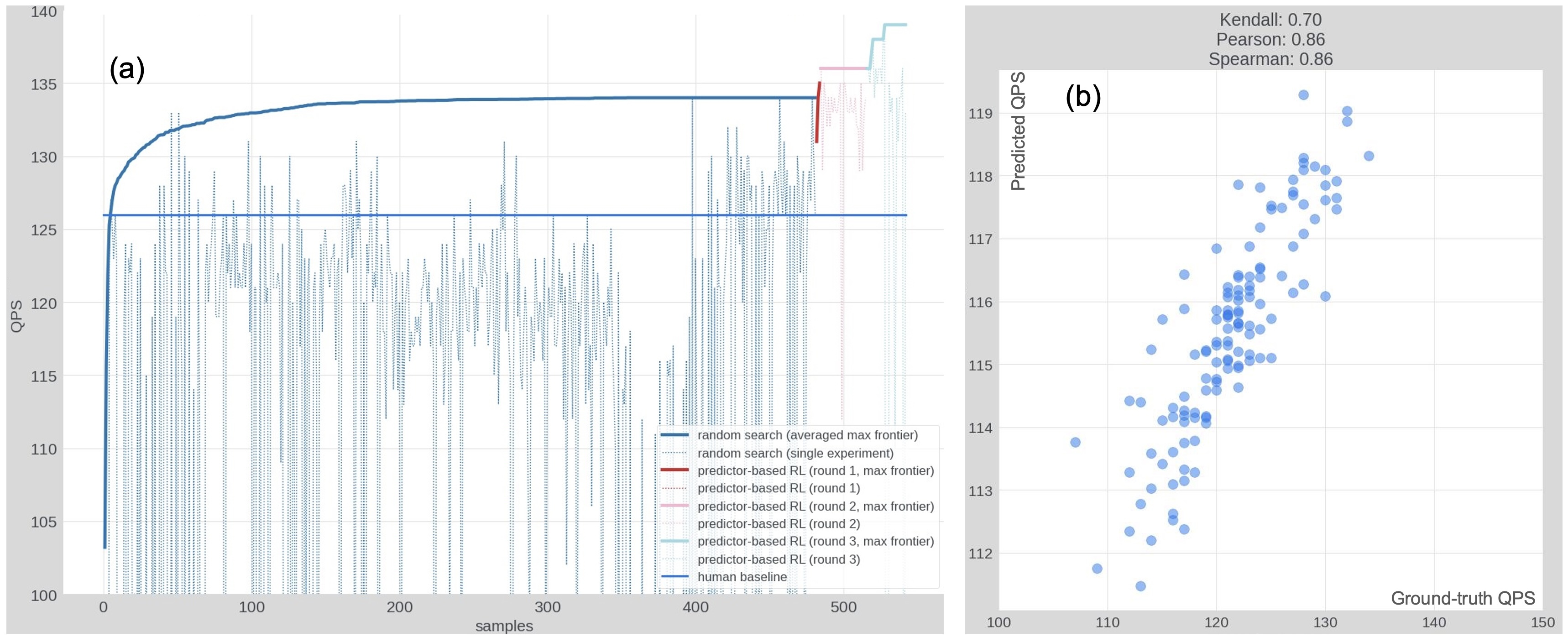}
\caption{(a) CubicML result (with predictor-based RL searcher) when optimizing QPS of an ads recommendation model by searching FSDP and other co-design hyper-parameters. x-axis: the number of configurations/jobs CubicML run. y-axis: $90$-th percentile QPS. In the plots, QPS values are normalized/divided by a constant. The same color illustrates the same round of search with dotted line indicating ground-truth QPS value per sample/configuration and solid line indicating maximal QPS frontier observed as each round proceeds. For random search round, we average maximal frontiers over $100$ perturbations. Note that in ``round 1'' predictor-based RL search, we only launch a few top jobs for a quick test during development, ending up a very short line. Jobs failed because of out-of-memory or infra failures are not plotted. (b) rank correlation of ground-truth QPS and predicted QPS by the ``predictor''. A validation dataset is used here. Note that we use pairwise ranking loss to train the predictor and the absolute values of predicted QPS does not need to approximate the ground truth QPS as long as the rank correlation is high.}
\label{fig:autoco_exp_fsdp_ads}
\end{figure}

ZeRO Sharding Optimization is a data parallelism to scale ML models. ZeRO~\cite{rajbhandari2020zero,rasley2020deepspeed} reduces memory usage per GPU by sharding optimizer states (stage 1), gradients (stage 2) and parameters (stage 3) across GPUs. More aggressive sharding (i.e. later stage) can save more memory but paying more cost of communication slowing down training.
We use Pytorch FSDP~\cite{zhao2023pytorch} implementation, a variant inspired by ZeRO, in our experiment.
Our training hardware is the Grand Teton platform~\footnote{https://engineering.fb.com/2022/10/18/open-source/ocp-summit-2022-grand-teton/} with 128 NVIDIA H100 GPUs (8 GPUs per node).
The recommendation model is an ads multi-task model for both click through rate (CTR) and impression conversion rate (CVR) prediction, adopting the backbone of Wukong~\cite{zhang2024wukong} with $11$ layers. The model has $73$ billion parameters excluding embeddings of category features.
To optimize training speed measured by example/query per second (QPS), we design a search space with $5 \times 3^{11} \times 10 \approx 8.9 \times 10 ^ 6$ configurations, covering layer-wise FSDP sharding strategy, batch size and more in Appendix~\ref{appendix:ads_space}.

We use CubicML to search a configuration to maximize QPS. To reduce QPS reading variance, we train each job for $2000$ global mini-batches and use $90$-th percentile as the QPS metric. The search result is plotted in Figure~\ref{fig:autoco_exp_fsdp_ads}~(a).
Appendix~\ref{appendix:cubicml_train_details} includes details on how we trained the predictor and RL agent in CubicML.
In Figure~\ref{fig:autoco_exp_fsdp_ads}~(a), CubicML first randomly sampled around $480$ jobs to test the limit of random search method, and then iterated three rounds of the predictor-based RL search as explained in Appendix~\ref{appendix:cubicml_train_details}.
In Figure~\ref{fig:autoco_exp_fsdp_ads}~(a), the random search performance saturates quickly which proves the challenge in tuning this search space; however, after random search, QPS jumps quickly in each new round which proves the QPS uplifting ability of the predictor-based RL search in our solution.
Compared with the human baseline tuned by in-house engineers, CubicML achieved $10.3\%$ QPS boost which is significant in industry-scale recommendation models with power saving of MegaWatts. 
Moreover, the whole process is automated without human tuning. 

In each predictor-based RL round, we rank configurations based on the predicted QPS and launch jobs with higher predicted QPS first. We find that the ``max frontier'' (in solid lines) of QPS jumps more often at the beginning and real-time QPS (in dotted lines) trends down (which is very obvious in the last round). Both observations demonstrate the accurate rank of configurations and the accuracy of the predictor.
The accuracy of the predictor is very essential in CubicML, because it models system performance and functions as the proxy that RL uses to propose top configurations.
We evaluate the rank correlation of the predictor in Figure~\ref{fig:autoco_exp_fsdp_ads}~(b).
The result shows that our predictor can accurately score the ground-truth QPS with Kendall Tau $0.7$, Pearson $0.86$ and Spearman $0.86$.

\subsection{ML Prediction of Distributed Training Performance of Large Language Models}

As discussed, accurate prediction of system performance is essential in CubicML.
We have proved its efficacy when optimizing ZeRO, but the search space can grow exponentially as we scale up large language models (LLM) in distributed systems.
In this experiment, we evaluate how accurate the predictor can predict the training speed (words/tokens per second) when training LLM~\cite{dubey2024llama}.

We use training speed WPS (words/tokens per second) as the performance metric that CubicML targets to model.
The search space includes configurations detailed in Appendix~\ref{appendix:llm_space}, including Transformer model architecture configurations~\cite{vaswani2017attention}, distributed training co-design configurations~\cite{dubey2024llama} and system infra setup.
Each configuration has a wide range of values.
For example, the number of GPUs can be as small as $8$ and up to $16,384$; the sequence length ranges between $2048$ and $131,072$; the number of parameters ranges from $27$ million up to $405$ billion.
The predictor training details are in Appendix~\ref{appendix:llm_predictor}.
We accumulated $568$ LLM jobs over time during LLM development.
We use $145$ examples as validation dataset and the rest ($423$ examples) as training data. The predicted WPS versus actual WPS is plotted in Figure~\ref{fig:llm_prediction}, where we performed different training-validation split to simulate three different real-world use cases:
\begin{itemize}
    \item using random split which simulates the use case where CubicML samples and profiles configurations in real-time (i.e. online) for optimization as model and systems evolve;
    \item using examples in older jobs as training dataset and validating the predictor by jobs launched later sorted by timestamps. Note that the timestamp gap between jobs can be as long as half a year. This simulates the use case where CubicML targets on reusing historical jobs to jump start search with the purpose of saving compute resources. It also tests the generalization of the predictor when the distributed ML systems evolve in the wild;
    \item using examples at smaller scales ($8 \leq  \#GPUs \leq 3072$) to predict WPS under larger scales ($4096 \leq  \#GPUs \leq 16384$). This may simulate the random split use case but only profile small scale jobs to save compute and predict performance  of large scale jobs. Note that this simulation may not fully fulfill its purpose because more smaller-scale jobs were launched earlier as shown in Figure~\ref{fig:gpu_timestamp}~(left) introducing unattempted bias to the second use case.
\end{itemize}

As shown in Figure~\ref{fig:llm_prediction}~(left), CubicML predictor achieves great rank correlation metrics of Kendall Tau $0.88$, Pearson $0.97$ and Spearman $0.97$, proving the applicability of CubicML to optimize distributed LLM training performance.
Figure~\ref{fig:llm_prediction}~(middle) shows correlation drop when generalizing from history data to future, which is expected because of the data distribution shift (caused by deprecation and update of models and systems in the development span); however, the rank correlation is still decent proving the value of reusing historical jobs in the wild to save compute resource when using CubicML.
Figure~\ref{fig:llm_prediction}~(right) shows more challenges if we only profile small scale jobs and attempt to generalize to large scale jobs. However, the rank correlation is still better than random guess which should regress all correlation metrics to $0.0$. We expect some real-time profiling jobs at large scale are required to correct the prediction when using CubicML in this use case.

Last but not least, we evaluate how many examples (i.e. the number of jobs we need to profile) when building the predictor in CubicML in Figure~\ref{fig:gpu_timestamp} (right). Random dataset split is used in this study. $50$ examples already provides a decent prediction performance to start, and Pearson and Spearman reach $\geq 0.9$ after $150$ examples. This implies that a relatively small amount of profiling jobs is enough for CubicML to search. Combined with the fact that the GPU cost of each profiling job is low, CubicML can be an efficient automated solution to optimize distributed LLM training performance.

%% file: appendix.tex
\newpage
\section{Appendix}
\subsection{Training details of the predictor and RL agent in CubicML}

\subsubsection{The Predictor and RL Agent when Optimizing Ads Recommendation Models}
\label{appendix:cubicml_train_details}

We train an ensemble of $10$ neural networks with identical architecture as a predictor. To train a neural network, each feature (i.e. a configuration) is encoded as one-hot and the neural network has a single hidden layer with $1600$ dimensions with dropout rate $0.5$.
The predictor is trained by the AMSGrad~\cite{reddi2019convergence} optimizer for $200$ epochs with batch size $64$, learning rate $0.001$ and weight decay $0.005$.
The margin is set to $0.001$ in Margin Ranking Loss.

In Figure~\ref{fig:autoco_exp_fsdp_ads}~(a), CubicML first randomly sample around $480$ jobs to test the limit of random search method, and then iterates three rounds of the predictor-based RL search as explained in Figure~\ref{fig:autoco}; that is, random search builds the initial historical job data which is later used to train the predictor to predict QPS of any configuration; then the RL searcher samples $2000$ configurations to maximize the reward (i.e. predicted QPS by the predictor). The configurations with top $50$ rewards are launched as new jobs which are added to historical job data upon completion to retrain the predictor for a next round of sampling. 

During a RL search, its agent parameters are optimized by Adam optimizer with learning rate $0.01$ and batch size $30$.
In our experiment, we found that a single RL search returned similar configurations around a local minimum, to resolve which, we rerun three RL search trials with different randomization seeds and select top configurations from all trials.

\subsubsection{The Predictor when Predicting Large Language Models Training Speed}
\label{appendix:llm_predictor}
To evaluate the ability to predict LLM training performance, we simply use gradient boosting regressor as the predictor.
Integer and floating-point configurations are encoded as numerical features, Boolean and object (e.g. string) configurations are encoded as categorical features, ending up with feature vector length of $117$.

\subsection{Search Spaces in CubicML Experiments}
\subsubsection{Ads Recommendation Models}
\label{appendix:ads_space}
To optimize training speed measured by example/query per second (QPS), we design a search space with $5 \times 3^{11} \times 10 \approx 8.9 \times 10 ^ 6$ configurations, generated from
\begin{itemize}
    \item FSDP sharding strategy \footnote{https://pytorch.org/docs/stable/fsdp.html\#torch.distributed.fsdp.ShardingStrategy} per layer: \texttt{FULL\_SHARD} (ZeRO stage 3), \texttt{SHARD\_GRAD\_OP} (ZeRO stage 2) and \texttt{NO\_SHARD} (standard distributed data parallelism)
    \item Local batch size per GPU: minimum is $1024$ and maximum is $1536$ with step size $128$. These batch sizes were verified with minor impact on model accuracy, so we can focus on QPS optimization by short-term profiling jobs without worrying about long-term accuracy.
    \item Storage reservation policy:  
    \begin{itemize}
    \item FixedPercentage: A fixed percentage of HBM reserved for dense paramemers and runtime memory, ranging from 0.77 to 0.85, each step is 0.01. 
    \item MemoryBalanced: The embedding sharding planner will find the minimal HBM used for emebdding tables, leave rest for dense parameters and runtime memory.
    \end{itemize}
\end{itemize}

\subsubsection{Large Language Models}
\label{appendix:llm_space}
We use training speed WPS (words/tokens per second) as the performance metric that CubicML targets to model.
The input features were concatenation of configurations (which influences WPS) below: 
\begin{itemize}
    \item Transformer model architecture configurations~\cite{vaswani2017attention}, such as the number of layers, the numbers of heads, model dimension, feed forward network dimensions, batch size, sequence length, etc
    \item Distributed training co-design configurations~\cite{dubey2024llama}, such as which parallelism strategies (tensor parallelism, pipeline parallelism, context parallelism, data parallelism) should be selected and their configurations (e.g. group sizes), precision formats (FP8 and BF16), etc
    \item Distributed system infrastructure info: the number of GPUs and hardware types
\end{itemize}
Each configuration has a wide range of values.
For example, the number of GPUs can be as small as $8$ and up to $16,384$; the sequence length ranges between $2048$ and $131,072$; the number of parameters ranges from $27$ million up to $405$ billion.

\subsection{Figures of Experiments for Large Language Models}

\begin{figure}[h]
\centering
\includegraphics[width=1.0\textwidth]{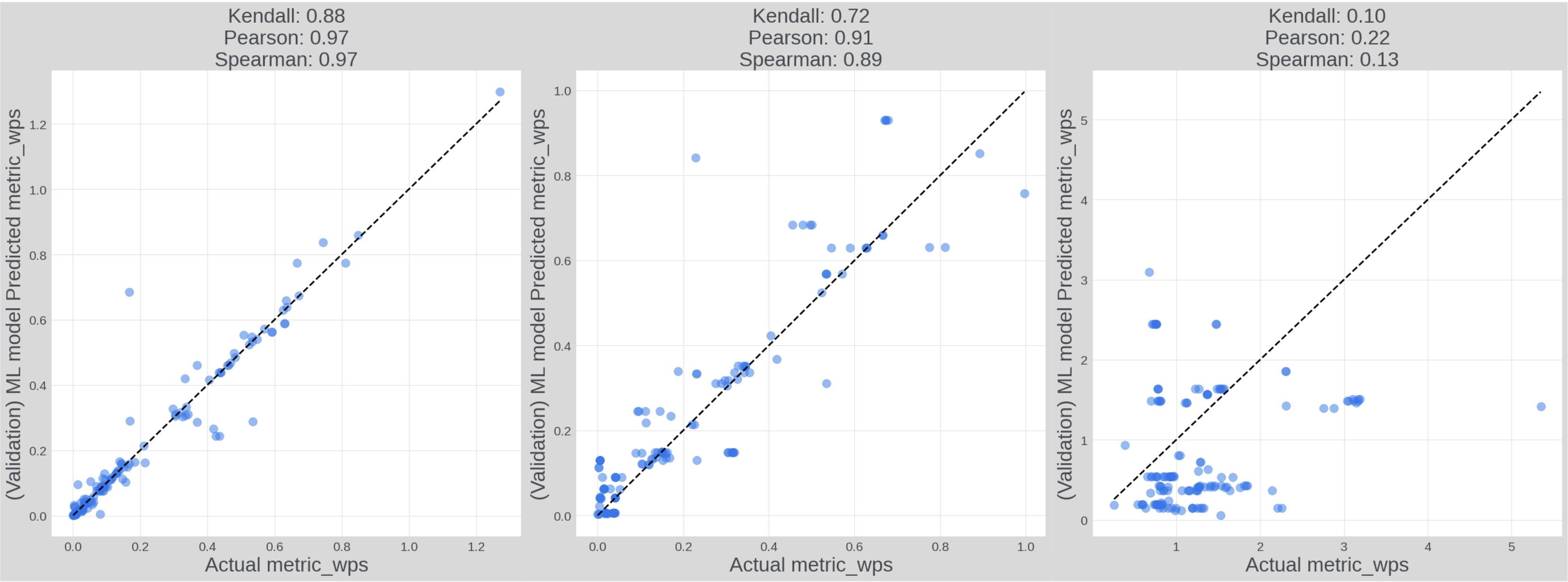}
\caption{Predicted WPS versus actual WPS by the predictor in CubicML. Left: random split of dataset; middle: use examples in older jobs to predict newer jobs; right: use examples with smaller numbers of GPUs to predict larger scale with more GPUs. Note that all WPS values are normalized/divided by a constant in a plot.}
\label{fig:llm_prediction}
\end{figure}

\begin{figure}[h]
\centering
\includegraphics[width=1.\textwidth]{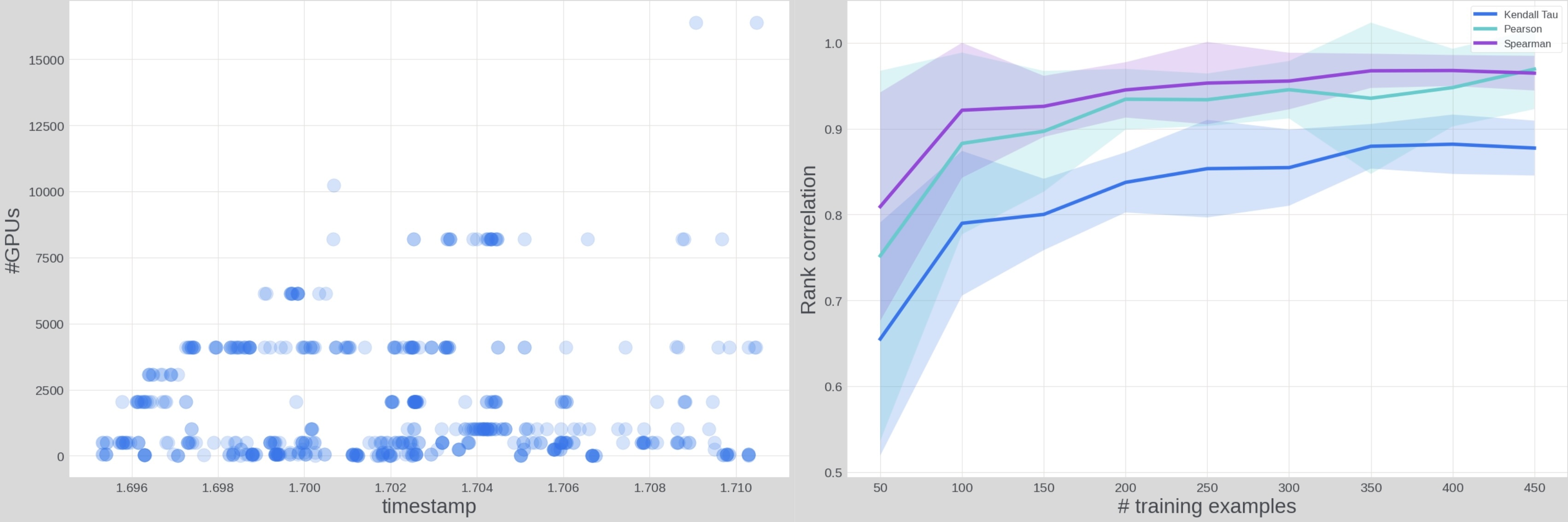}
\caption{Left: the number of GPUs used in jobs sorted by timestamps. Note that timestamps are normalized/divided by a constant. Right: rank correlation versus the number of training examples by random split of dataset. Shading bands are $\pm 2.0 \times$ standard deviation over ten random perturbations.}
\label{fig:gpu_timestamp}
\end{figure}

%% file: neurips_2024.bbl
\begin{thebibliography}{22}
\providecommand{\natexlab}[1]{#1}
\providecommand{\url}[1]{\texttt{#1}}
\expandafter\ifx\csname urlstyle\endcsname\relax
  \providecommand{\doi}[1]{doi: #1}\else
  \providecommand{\doi}{doi: \begingroup \urlstyle{rm}\Url}\fi

\bibitem[Kaplan et~al.(2020)Kaplan, McCandlish, Henighan, Brown, Chess, Child,
  Gray, Radford, Wu, and Amodei]{kaplan2020scaling}
Jared Kaplan, Sam McCandlish, Tom Henighan, Tom~B Brown, Benjamin Chess, Rewon
  Child, Scott Gray, Alec Radford, Jeffrey Wu, and Dario Amodei.
\newblock Scaling laws for neural language models.
\newblock \emph{arXiv preprint arXiv:2001.08361}, 2020.

\bibitem[Peebles and Xie(2023)]{peebles2023scalable}
William Peebles and Saining Xie.
\newblock Scalable diffusion models with transformers.
\newblock In \emph{Proceedings of the IEEE/CVF International Conference on
  Computer Vision}, pages 4195--4205, 2023.

\bibitem[Zhang et~al.(2024)Zhang, Luo, Chen, Nie, Liu, Guo, Zhao, Li, Hao, Yao,
  et~al.]{zhang2024wukong}
Buyun Zhang, Liang Luo, Yuxin Chen, Jade Nie, Xi~Liu, Daifeng Guo, Yanli Zhao,
  Shen Li, Yuchen Hao, Yantao Yao, et~al.
\newblock Wukong: Towards a scaling law for large-scale recommendation.
\newblock \emph{arXiv preprint arXiv:2403.02545}, 2024.

\bibitem[Anil et~al.(2022)Anil, Gadanho, Huang, Jacob, Li, Lin, Phillips, Pop,
  Regan, Shamir, et~al.]{anil2022factory}
Rohan Anil, Sandra Gadanho, Da~Huang, Nijith Jacob, Zhuoshu Li, Dong Lin, Todd
  Phillips, Cristina Pop, Kevin Regan, Gil~I Shamir, et~al.
\newblock On the factory floor: Ml engineering for industrial-scale ads
  recommendation models.
\newblock \emph{arXiv preprint arXiv:2209.05310}, 2022.

\bibitem[Ardalani et~al.(2022)Ardalani, Wu, Chen, Bhushanam, and
  Aziz]{ardalani2022understanding}
Newsha Ardalani, Carole-Jean Wu, Zeliang Chen, Bhargav Bhushanam, and Adnan
  Aziz.
\newblock Understanding scaling laws for recommendation models.
\newblock \emph{arXiv preprint arXiv:2208.08489}, 2022.

\bibitem[Rajbhandari et~al.(2020)Rajbhandari, Rasley, Ruwase, and
  He]{rajbhandari2020zero}
Samyam Rajbhandari, Jeff Rasley, Olatunji Ruwase, and Yuxiong He.
\newblock Zero: Memory optimizations toward training trillion parameter models.
\newblock In \emph{SC20: International Conference for High Performance
  Computing, Networking, Storage and Analysis}, pages 1--16. IEEE, 2020.

\bibitem[Rasley et~al.(2020)Rasley, Rajbhandari, Ruwase, and
  He]{rasley2020deepspeed}
Jeff Rasley, Samyam Rajbhandari, Olatunji Ruwase, and Yuxiong He.
\newblock Deepspeed: System optimizations enable training deep learning models
  with over 100 billion parameters.
\newblock In \emph{Proceedings of the 26th ACM SIGKDD International Conference
  on Knowledge Discovery \& Data Mining}, pages 3505--3506, 2020.

\bibitem[Zhao et~al.(2023)Zhao, Gu, Varma, Luo, Huang, Xu, Wright, Shojanazeri,
  Ott, Shleifer, et~al.]{zhao2023pytorch}
Yanli Zhao, Andrew Gu, Rohan Varma, Liang Luo, Chien-Chin Huang, Min Xu, Less
  Wright, Hamid Shojanazeri, Myle Ott, Sam Shleifer, et~al.
\newblock Pytorch fsdp: experiences on scaling fully sharded data parallel.
\newblock \emph{arXiv preprint arXiv:2304.11277}, 2023.

\bibitem[Shoeybi et~al.(2019)Shoeybi, Patwary, Puri, LeGresley, Casper, and
  Catanzaro]{shoeybi2019megatron}
Mohammad Shoeybi, Mostofa Patwary, Raul Puri, Patrick LeGresley, Jared Casper,
  and Bryan Catanzaro.
\newblock Megatron-lm: Training multi-billion parameter language models using
  model parallelism.
\newblock \emph{arXiv preprint arXiv:1909.08053}, 2019.

\bibitem[Narayanan et~al.(2021)Narayanan, Shoeybi, Casper, LeGresley, Patwary,
  Korthikanti, Vainbrand, Kashinkunti, Bernauer, Catanzaro,
  et~al.]{narayanan2021efficient}
Deepak Narayanan, Mohammad Shoeybi, Jared Casper, Patrick LeGresley, Mostofa
  Patwary, Vijay Korthikanti, Dmitri Vainbrand, Prethvi Kashinkunti, Julie
  Bernauer, Bryan Catanzaro, et~al.
\newblock Efficient large-scale language model training on gpu clusters using
  megatron-lm.
\newblock In \emph{Proceedings of the International Conference for High
  Performance Computing, Networking, Storage and Analysis}, pages 1--15, 2021.

\bibitem[Wen et~al.(2017)Wen, Xu, Yan, Wu, Wang, Chen, and Li]{wen2017terngrad}
Wei Wen, Cong Xu, Feng Yan, Chunpeng Wu, Yandan Wang, Yiran Chen, and Hai Li.
\newblock Terngrad: Ternary gradients to reduce communication in distributed
  deep learning.
\newblock \emph{Advances in neural information processing systems}, 30, 2017.

\bibitem[Peng et~al.(2023)Peng, Wu, Wei, Zhao, Yang, Liu, Xiong, Yang, Ni, Hu,
  et~al.]{peng2023fp8}
Houwen Peng, Kan Wu, Yixuan Wei, Guoshuai Zhao, Yuxiang Yang, Ze~Liu, Yifan
  Xiong, Ziyue Yang, Bolin Ni, Jingcheng Hu, et~al.
\newblock Fp8-lm: Training fp8 large language models.
\newblock \emph{arXiv preprint arXiv:2310.18313}, 2023.

\bibitem[He et~al.(2021)He, Zhao, and Chu]{he2021automl}
Xin He, Kaiyong Zhao, and Xiaowen Chu.
\newblock Automl: A survey of the state-of-the-art.
\newblock \emph{Knowledge-based systems}, 212:\penalty0 106622, 2021.

\bibitem[Zheng et~al.(2022)Zheng, Li, Zhang, Zhuang, Chen, Huang, Wang, Xu,
  Zhuo, Xing, et~al.]{zheng2022alpa}
Lianmin Zheng, Zhuohan Li, Hao Zhang, Yonghao Zhuang, Zhifeng Chen, Yanping
  Huang, Yida Wang, Yuanzhong Xu, Danyang Zhuo, Eric~P Xing, et~al.
\newblock Alpa: Automating inter-and $\{$Intra-Operator$\}$ parallelism for
  distributed deep learning.
\newblock In \emph{16th USENIX Symposium on Operating Systems Design and
  Implementation (OSDI 22)}, pages 559--578, 2022.

\bibitem[Zhang et~al.(2023)Zhang, Diao, Wang, Cao, Gu, Si, Shi, Zheng, Wu, and
  Lin]{zhang2023auto}
Shiwei Zhang, Lansong Diao, Siyu Wang, Zongyan Cao, Yiliang Gu, Chang Si, Ziji
  Shi, Zhen Zheng, Chuan Wu, and Wei Lin.
\newblock Auto-parallelizing large models with rhino: A systematic approach on
  production ai platform.
\newblock \emph{arXiv preprint arXiv:2302.08141}, 2023.

\bibitem[Lin et~al.(2023)Lin, Wu, Li, Li, and Li]{lin2023uniap}
Hao Lin, Ke~Wu, Jie Li, Jun Li, and Wu-Jun Li.
\newblock Uniap: Unifying inter-and intra-layer automatic parallelism by mixed
  integer quadratic programming.
\newblock \emph{arXiv preprint arXiv:2307.16375}, 2023.

\bibitem[Chen et~al.(2018)Chen, Moreau, Jiang, Zheng, Yan, Shen, Cowan, Wang,
  Hu, Ceze, et~al.]{chen2018tvm}
Tianqi Chen, Thierry Moreau, Ziheng Jiang, Lianmin Zheng, Eddie Yan, Haichen
  Shen, Meghan Cowan, Leyuan Wang, Yuwei Hu, Luis Ceze, et~al.
\newblock $\{$TVM$\}$: An automated $\{$End-to-End$\}$ optimizing compiler for
  deep learning.
\newblock In \emph{13th USENIX Symposium on Operating Systems Design and
  Implementation (OSDI 18)}, pages 578--594, 2018.

\bibitem[Yin et~al.(2024)Yin, Liu, Sun, Chen, Zhang, Liu, Sehgal, Panchal,
  Hotaj, Liu, et~al.]{yin2024automl}
Hang Yin, Kuang-Hung Liu, Mengying Sun, Yuxin Chen, Buyun Zhang, Jiang Liu,
  Vivek Sehgal, Rudresh~Rajnikant Panchal, Eugen Hotaj, Xi~Liu, et~al.
\newblock Automl for large capacity modeling of meta's ranking systems.
\newblock In \emph{Companion Proceedings of the ACM on Web Conference 2024},
  pages 374--382, 2024.

\bibitem[Wen et~al.(2020)Wen, Liu, Chen, Li, Bender, and
  Kindermans]{wen2020neural}
Wei Wen, Hanxiao Liu, Yiran Chen, Hai Li, Gabriel Bender, and Pieter-Jan
  Kindermans.
\newblock Neural predictor for neural architecture search.
\newblock In \emph{European Conference on Computer Vision}, pages 660--676.
  Springer, 2020.

\bibitem[Dubey et~al.(2024)Dubey, Jauhri, Pandey, Kadian, Al-Dahle, Letman,
  Mathur, Schelten, Yang, Fan, et~al.]{dubey2024llama}
Abhimanyu Dubey, Abhinav Jauhri, Abhinav Pandey, Abhishek Kadian, Ahmad
  Al-Dahle, Aiesha Letman, Akhil Mathur, Alan Schelten, Amy Yang, Angela Fan,
  et~al.
\newblock The llama 3 herd of models.
\newblock \emph{arXiv preprint arXiv:2407.21783}, 2024.

\bibitem[Vaswani et~al.(2017)Vaswani, Shazeer, Parmar, Uszkoreit, Jones, Gomez,
  Kaiser, and Polosukhin]{vaswani2017attention}
Ashish Vaswani, Noam Shazeer, Niki Parmar, Jakob Uszkoreit, Llion Jones,
  Aidan~N Gomez, \L~ukasz Kaiser, and Illia Polosukhin.
\newblock Attention is all you need.
\newblock In I.~Guyon, U.~Von Luxburg, S.~Bengio, H.~Wallach, R.~Fergus,
  S.~Vishwanathan, and R.~Garnett, editors, \emph{Advances in Neural
  Information Processing Systems}, volume~30. Curran Associates, Inc., 2017.

\bibitem[Reddi et~al.(2019)Reddi, Kale, and Kumar]{reddi2019convergence}
Sashank~J Reddi, Satyen Kale, and Sanjiv Kumar.
\newblock On the convergence of adam and beyond.
\newblock \emph{arXiv preprint arXiv:1904.09237}, 2019.

\end{thebibliography}
